\documentclass[a4paper, 10 pt, conference]{IEEEtran}  

\IEEEoverridecommandlockouts

\usepackage{graphicx}
\usepackage{cite}
\usepackage{amssymb}
\usepackage{bm}
\usepackage{tcolorbox}
\usepackage{amsmath}
\usepackage{tabularray}
\usepackage[super]{nth}
\usepackage{color,soul}

\title{\LARGE \bf
Odometry Calibration and Pose Estimation of a 4WIS4WID Mobile Wall Climbing Robot
}

\author{Branimir Ćaran$^{1}$, Vladimir Milić$^{1}$, Marko Švaco$^{1}$ and Bojan Jerbić$^{1,2}$
\thanks{$^{1}$Authors are with Faculty of Mechanical Engineering and Naval Arhitecture, University of Zagreb, Zagreb, 10000, Croatia
        {\tt\small branimir.caran@fsb.unizg.hr}}
\thanks{$^{2}$Bojan Jerbić is with Croatian Academy of Sciences and Arts, Zagreb, 10000, Croatia}
}

\usepackage{fancyhdr}
\begin{document}
\maketitle
\thispagestyle{fancy}   
\pagestyle{fancy}       

\begin{abstract}
This paper presents the design of a pose estimator for a four wheel independent steer four wheel independent drive (4WIS4WID) wall climbing mobile robot, based on the fusion of multimodal measurements, including wheel odometry, visual odometry, and an inertial measurement unit (IMU) data using Extended Kalman Filter (EKF) and Unscented Kalman Filter (UKF). The pose estimator is a critical component of wall climbing mobile robots, as their operational environment involves carrying precise measurement equipment and maintenance tools in construction, requiring information about pose on the building at the time of measurement. Due to the complex geometry and material properties of building façades, the use of traditional localization sensors such as laser, ultrasonic, or radar is often infeasible for wall-climbing robots. Moreover, GPS-based localization is generally unreliable in these environments because of signal degradation caused by reinforced concrete and electromagnetic interference. Consequently, robot odometry remains the primary source of velocity and position information, despite being susceptible to drift caused by both systematic and non-systematic errors. The calibrations of the robot’s systematic parameters were conducted using nonlinear optimization and Levenberg-Marquardt methods as Newton-Gauss and gradient-based model fitting methods, while Genetic algorithm and Particle swarm were used as stochastic based methods for kinematic parameter calibration. Performance and results of the calibration methods and pose estimators  were validated in detail with experiments on the experimental mobile wall climbing robot.
\end{abstract}

\renewcommand\IEEEkeywordsname{Keywords}
\begin{IEEEkeywords}
\textit{wall climbing robot,  four-wheel independent steer four-wheel independent drive mobile robot, odometry calibration, pose estimation, Extended Kalman Filter, Unscented Kalman Filter}
\end{IEEEkeywords}

\section{INTRODUCTION}
Mobile wall climbing robots for vertical surfaces belong to the category of mobile robots that are able to operate on vertical surfaces in order to perform specific tasks that usual ground mobile robot cannot\cite{NANSAI20163}. This type of mobile robot is capable of performing inspection, cleaning and maintenance tasks when access to the location of the work is not available or dangerous \cite{SCHMIDT20131288}. The basic division of robots for vertical surfaces is defined with respect to motion systems and methods of making contact with the vertical surface \cite{FANG202347}. Robots for vertical surfaces are divided into robots that move using wheels, tracks, walking robots, and hybrid robots that combine at least two of the previously mentioned locomotion methods. To achieve contact on vertical surfaces, robots use magnetic, pneumatic, mechanical, chemical, electrostatic, and hybrid adhesion systems \cite{FANG202219}. In research presented in this paper, hybrid mobile robots that use the principle of negative pressure and thrust systems such as quadrotors to compensate for gravitational effects are of basic interest. Due to the design of the robot and the placement of the adhesion system and thrust propellers, it is necessary to maintain constant orientation of the robot, therefore omnidirectional motion is preferred in such cases. To achieve omnidirectional movement, kinematic structure of four independently driven \cite{CHOKSHI2021} wheels and four independently steerable wheels is used \cite{LI20142015}.

For successful navigation, a mobile robot must be able to estimate its current position and, based on its current and desired position, define the control outputs in order to track the desired trajectory. Localization is a segment of mobile robotics that is of the greatest importance and where the most significant advancements have been made \cite{ULLAH2024100651}, which is not the case with mobile robots for vertical surfaces \cite{TAO202310}. Most indoor mobile robots rely on Light Detection and Ranging (LIDAR) sensors to perform simultaneous localization and mapping (SLAM) \cite{YAROVOI2024105344}. The enclosure and variability of indoor spaces allow for the detection of 2D features and localization in space, which, combined with the robot's odometry, enables precise localization. In addition to 2D LIDAR sensors, 3D LIDAR sensors are increasingly being used, which can provide more information about the space but are often more computationally demanding and slower in estimation. The use of optical sensors in outdoor environments is challenging due to varying lighting conditions and environmental factors, so the application of the Global Positioning System (GPS) becomes necessary if the environment permits. Due to its lower frequency (1 Hz), relying solely on GPS is not sufficient, which is why slow GPS is often combined with odometry and Inertial Measurement Unit (IMU) sensors to better localize the robot in space \cite{YOUSUF2021250}.\\
The use of LIDAR sensors and GPS in the case of climbing robots for vertical surfaces is not sufficient because the concrete infrastructure is located outdoors and near the reinforcement, which hinders the operation of the aforementioned sensors. In the literature, several new approaches to localization of wall climbing robots are emerging, in which external measuring devices are used to try to localize the robot.

In \cite{ZHOU202098} they used external camera with convolutional neural networks to localize mobile wall climbing robot. They tested their positioning scheme on multiple videos of the wall-climbing robot but did not provide the whole control scheme and use that as part of the feedback control loop. These results on the speed and accuracy of this positioning system clearly indicate its efficacy and potential for application in a variety of industrial settings but main disadvantage is need for external camera as well as intrinsic and extrinsic camera calibration. Zhang et al. in \cite{ZHANG2025} also used external depth camera fused together with IMU to localize mobile wall climbing robot on cilindrical shape with magnetic wheels. The results obtained prove their applicability for the autonomous localization in low-texture and magnetically disturbed environments. Yoo et al. \cite{YOO20211754} introduce a novel dual rope winch robot with two degrees of freedom and an integrated position tracking control system. It is designed for easy installation at both corners of the building using two synthetic ropes, allowing it to move freely along the wall surface while performing maintenance tasks. Their focus was on the characteristics and modeling of the ropes to estimate the robot’s position on the wall. The drawback of this method is that it requires placing the ropes on the edges of the wall, which is often not possible or may require climbing, which is not safe.

This paper considers the problem of pose estimation of the 4WIS4WID wall climbing mobile robot based on odometry, IMU and visual odometry from mobile robot. The goal is to calibrate mobile robot odometry in horizontal plane to get corrected parameters for the kinematic model, as the state estimator is based on odometry and to the best of author knowledge odometry calibration for 4WIS4WID mobile robot hasn't been done yet, develop and compare two pose estimators based on Extended Kalman Filter (EKF) and Unscented Kalman Filter (UKF) and compare them with ground truth values using \textit{OptiTrack} motion capture system. Two main contributions in this paper are 4WIS4WID mobile robot kinematic calibration and pose estimator that should be general for all wall climbing robot, no matter the robot structure or surface. According to that, paper is organized as follows. Used Wall climbing robot is described in Section \ref{sec_wcr}. Section \ref{sec_4wis4wid} describes odometry and odometry calibration for 4WIS4WID mobile robot. Pose estimator and experimental results were given in Sections \ref{sec_pose_estimator} and \ref{sec_experimental_validation} while Section \ref{sec_conclusion} gives final conclusions and future work.

\section{WALL CLIMBING ROBOT} \label{sec_wcr}
\begin{figure}[!ht]
\centering
\includegraphics[scale=0.4]{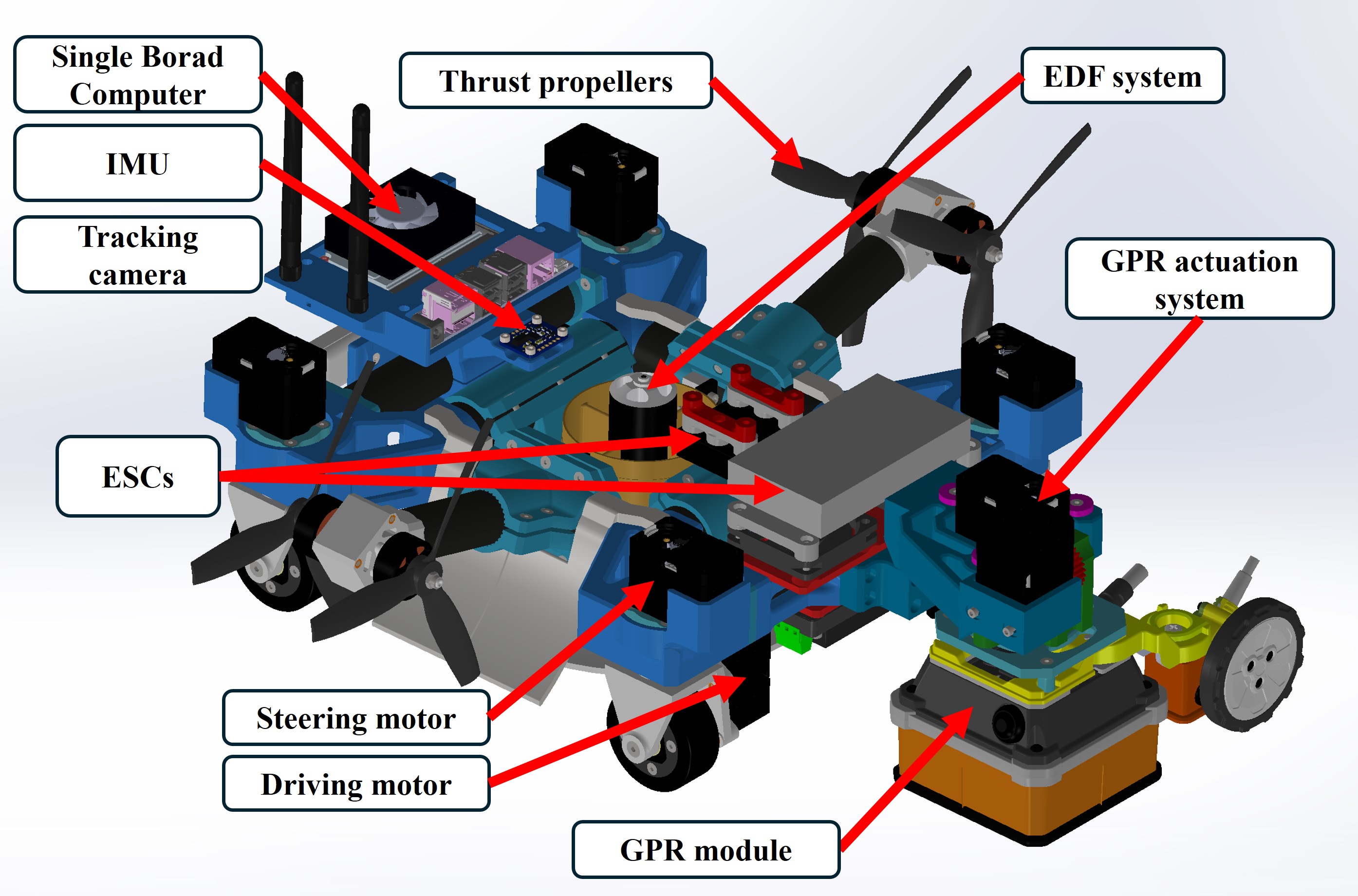}
\caption{Computer aided design of the wall robot.}\label{wcr_cad}
\end{figure}

\begin{figure}[!ht]
\centering
    \includegraphics[scale=0.11]{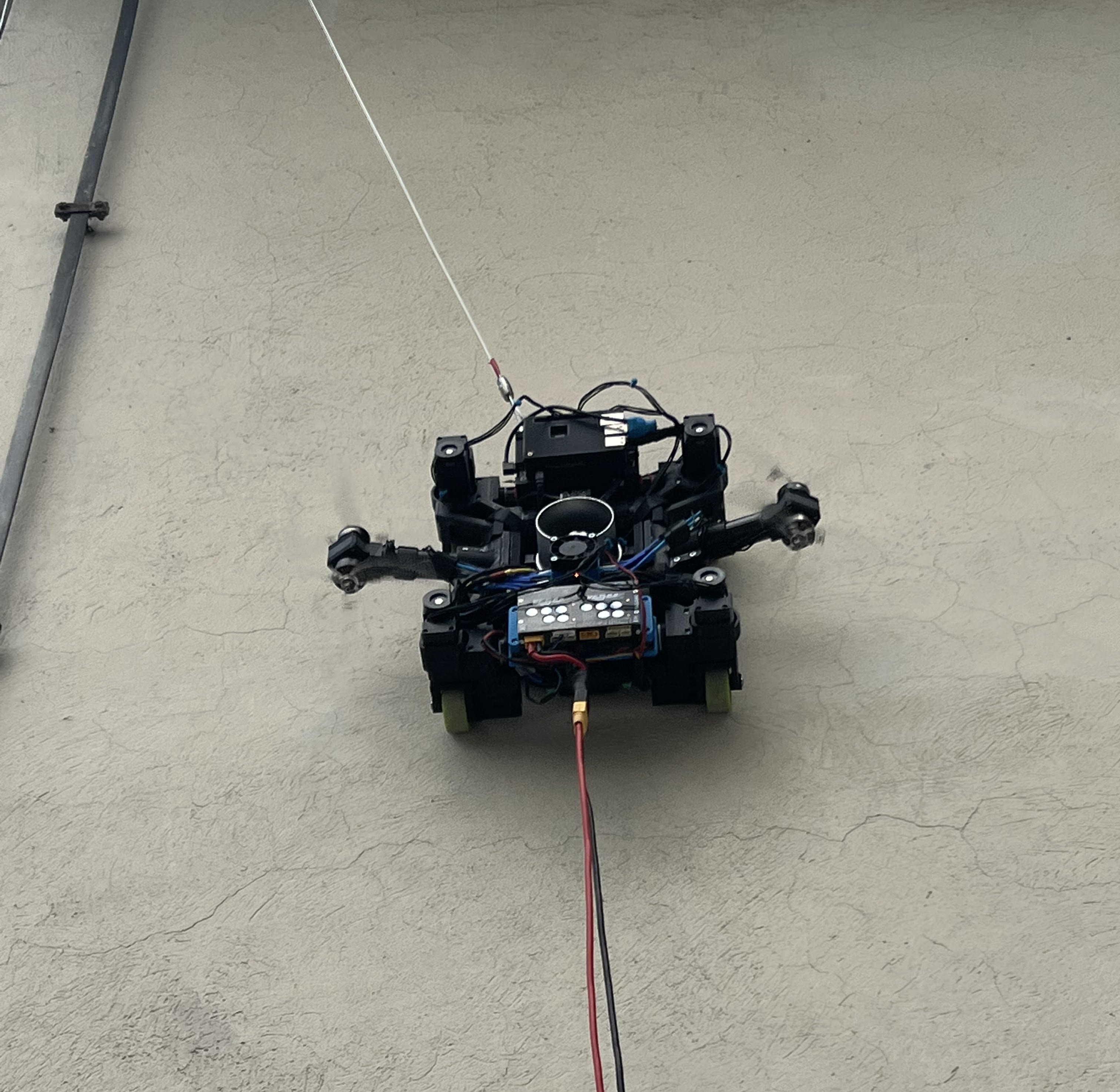} 
\caption{4WIS4WID wall climbing robot driving on vertical surface.}\label{fig_experimental_setup}
\end{figure}

In this paper, we used a custom-built mobile robot designed as a wall climbing robot capable of carrying non-destructive testing (NDT) equipment such as ground penetrating radar (GPR) or half cell potential (HCP) measurement devices. Figure \ref{wcr_cad} presents a computer-aided design model of the wall climbing robot, highlighting a dedicated mounting section for the GPR module.
For safety reasons during the experiments, the GPR was not used on the robot, while the robot payload analysis was performed in \cite{BOZIC2021MIPRO} and \cite{BOZIC022NDT}. Figure \ref{fig_experimental_setup} illustrates the experimental setup of the wall climbing robot. The robot chassis is made primarily from acrylonitrile styrene acrylate (ASA) using a 3D printing process, with selected components fabricated from carbon fiber tubes. The robot dimensions are 380×300 mm, has a total mass of 3.25 kg, and supports a payload of 1.5 kg. Its wheels are actuated and steered by Dynamixel XH430-W210-T smart servo motors, with rotational positions monitored via AMS AS5045 magnetic rotary sensors. The position and velocity control systems for steering and driving motors are designed and verified in \cite{CARAN2024}. In addition, the robot is equipped with a Bosch BNO055 inertial measurement unit (IMU) and Intel RealSense T265 tracking camera for visual odometry.
The software architecture centers on the NVIDIA Jetson NX, running ROS2. It manages motor commands for steering and speed while integrating different sensor data from other sensor for position estimation.

\section{4WIS4WID ODOMETRY} \label{sec_4wis4wid}
\begin{figure}[!ht]
\centering
\includegraphics[scale=0.8]{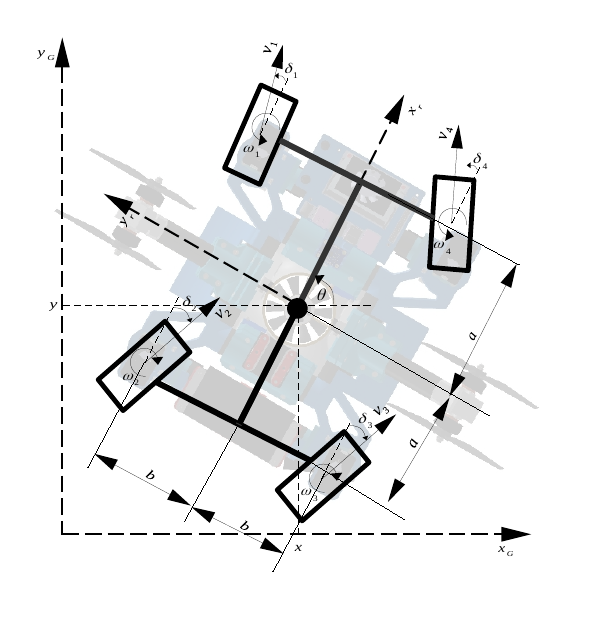}
\caption{4WIS4WID mobile robot kinematic structure.}\label{fig_rob_scheme}
\end{figure}

Wall movement is achieved through hybrid control, with the electric ducted fan (EDF) providing adhesion and thrust propellers counteracting the mass. The 4WIS4WID kinematic structure ensures the robot remains upright to maximize thrust propeller efficiency. The 4WIS4WID mobile robot has the ability to steer and drive each of the four wheels. As shown in Figure \ref{fig_rob_scheme}, rigid body constraints for each wheel velocity is presented as
\begin{eqnarray} 
 v_{xi} &=& v_i\cos{(\delta_i)} = v_x - y_{wi}\omega \nonumber,\\ 
 v_{yi} &=& v_i\sin{(\delta_i)} = v_y + x_{wi}\omega \label{eq_vxy},
\end{eqnarray}
where $v_{xi}$ and $v_{yi}$ are velocity components of each i\textsuperscript{th} wheel ($i = 1, 2, 3, 4$), $v_i=\sqrt{v_x^2 + v_y^2}$ and $\delta_i$ are, respectively, linear velocity and steering angle of each wheel, $x_{wi}$ and $y_{wi}$ are wheel positions in robot coordinate frame.
Equations \ref{eq_vxy} with robot parameters are represented as
\begin{equation}
\underbrace{
\begin{bmatrix}
    1 & 0 & -b\\
    0 & 1 & a\\
    1 & 0 & -b\\
    0 & 1 & -a\\
    1 & 0 & b\\
    0 & 1 & -a\\
    1 & 0 & b\\
    0 & 1 & a
\end{bmatrix}
}_{\mathbf{P}}
\begin{bmatrix}
    v_x\\
    v_y\\
    \omega
\end{bmatrix}
=
\underbrace{
\begin{bmatrix}
    \text{c}(\delta_1) & 0 & 0 & 0\\
    \text{s}(\delta_1) & 0 & 0 & 0\\
    0 & \text{c}(\delta_2) & 0 & 0\\
    0 & \text{s}(\delta_2) & 0 & 0\\
    0 & 0 & \text{c}(\delta_3) & 0\\
    0 & 0 & \text{s}(\delta_3) & 0\\
    0 & 0 & 0 & \text{c}(\delta_4)\\
    0 & 0 & 0 & \text{s}(\delta_4)
\end{bmatrix}
}_{\mathbf{R}}
\underbrace{
\begin{bmatrix}
    v_1\\
    v_2\\
    v_3\\
    v_4
\end{bmatrix}
}_{\mathbf{v}}\label{eq_PR}
\end{equation} 
where $\text{c(x)} = \cos{(x)}$, $\text{s(x)} = \sin{(x)}$, By premultiplying (\ref{eq_PR}) with Moore–Penrose inverse $\mathbf{P}^\dag$  relationship between wheel velocities, steering angles, and the robot's velocities:
\begin{align}
    \begin{bmatrix}
        v_x\\
        v_y\\
        \omega
    \end{bmatrix}
    =
    \begin{bmatrix}
        \frac{\text{c}(\delta_1)}{4} & \frac{\text{c}(\delta_2)}{4} & \frac{\text{c}(\delta_3)}{4}& \frac{\text{c}(\delta_4)}{4}\\
        \frac{\text{s}(\delta_1)}{4} & \frac{\text{s}(\delta_2)}{4} & \frac{\text{s}(\delta_3)}{4} & \frac{\text{s}(\delta_4)}{4}\\
        K_1 & K_2 & K_3 & K_4
    \end{bmatrix}
    \begin{bmatrix}
        v_1\\
        v_2\\
        v_3\\
        v_4
    \end{bmatrix}
\end{align}
where $K_i = \frac{-y_{wi}\cos{(\delta_i)} + x_{wi}\sin{\delta_i}}{4x_{wi}^2 + 4y_{wi}^2}$.\\
The kinematics model of the 4WIS4WID mobile robot can be presented with 11 states as:
\begin{align}
    \begin{bmatrix}
        \dot{x}\\
        \dot{y}\\
        \dot{\theta}\\
        \dot{\varphi}_1\\
        \dot{\varphi}_2\\
        \dot{\varphi}_3\\
        \dot{\varphi}_4\\
        \dot{\delta}_1\\
        \dot{\delta}_2\\
        \dot{\delta}_3\\
        \dot{\delta}_4\\
    \end{bmatrix}
    =
    \begin{bmatrix}
        \frac{\text{c}_1}{4} & \frac{\text{c}_2}{4} & \frac{\text{c}_3}{4} & \frac{\text{c}_4}{4} & 0 & 0 & 0 & 0\\
        \frac{\text{s}_1}{4} & \frac{\text{s}_2}{4} & \frac{\text{s}_3}{4} & \frac{\text{s}_4}{4} & 0 & 0 & 0 & 0\\
        K_1 & K_2 & K_3 & K_4 & 0 & 0 & 0 & 0\\
        r_1^{-1} & 0 & 0 & 0 & 0 & 0 & 0 & 0\\
        0 & r_2^{-1} & 0 & 0 & 0 & 0 & 0 & 0\\
        0 & 0 & r_3^{-1} & 0 & 0 & 0 & 0 & 0\\
        0 & 0 & 0 & r_4^{-1} & 0 & 0 & 0 & 0\\
        0 & 0 & 0 & 0 & 1 & 0 & 0 & 0\\
        0 & 0 & 0 & 0 & 0 & 1 & 0 & 0\\
        0 & 0 & 0 & 0 & 0 & 0 & 1 & 0\\
        0 & 0 & 0 & 0 & 0 & 0 & 0 & 1\\
    \end{bmatrix}
    \begin{bmatrix}
        v_1\\
        v_2\\
        v_3\\
        v_4\\
        \omega_1\\
        \omega_2\\
        \omega_3\\
        \omega_4
    \end{bmatrix} \label{eq_kinematics}
\end{align}
where $\text{c}_i=\cos{(\delta_i + \theta)}$, $\text{s}_i=\sin{(\delta_i + \theta)}$ and $r_i$ is wheel radius of each wheel. From (\ref{eq_kinematics}) it can been seen that mechanical parameters of the robot $x_{wi}$, $y_{wi}$, which are correlated with robot's dimensions $a$ and $b$, and wheel radius $r_i$ have influence on robot's odometry. Similar kinematic model is derived in \cite{LI20142015}.

\subsection{Initial odometry tests}

\begin{figure}[!ht]
\begin{center}
    \includegraphics[scale=0.50]{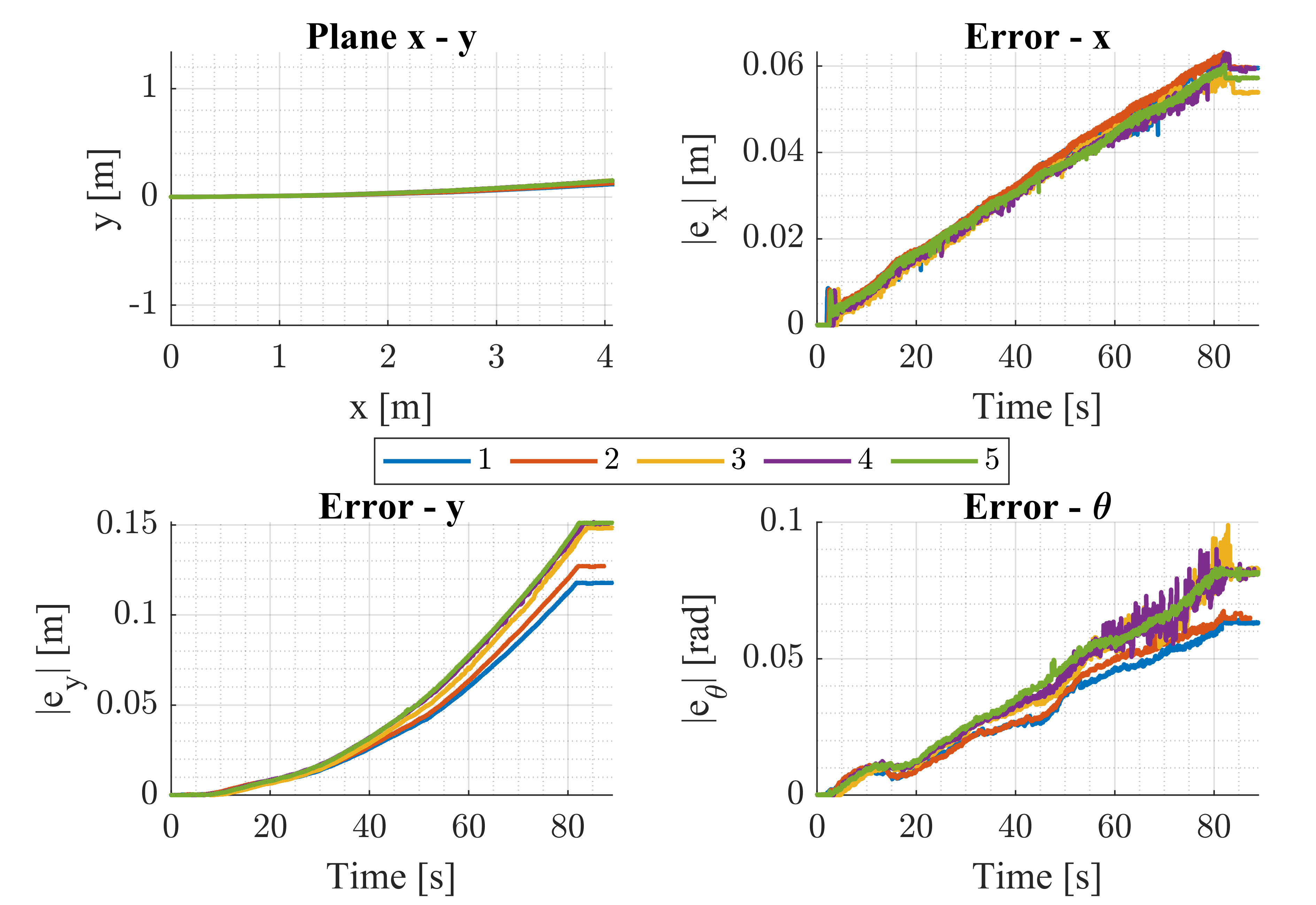} 
\includegraphics[scale=0.50]{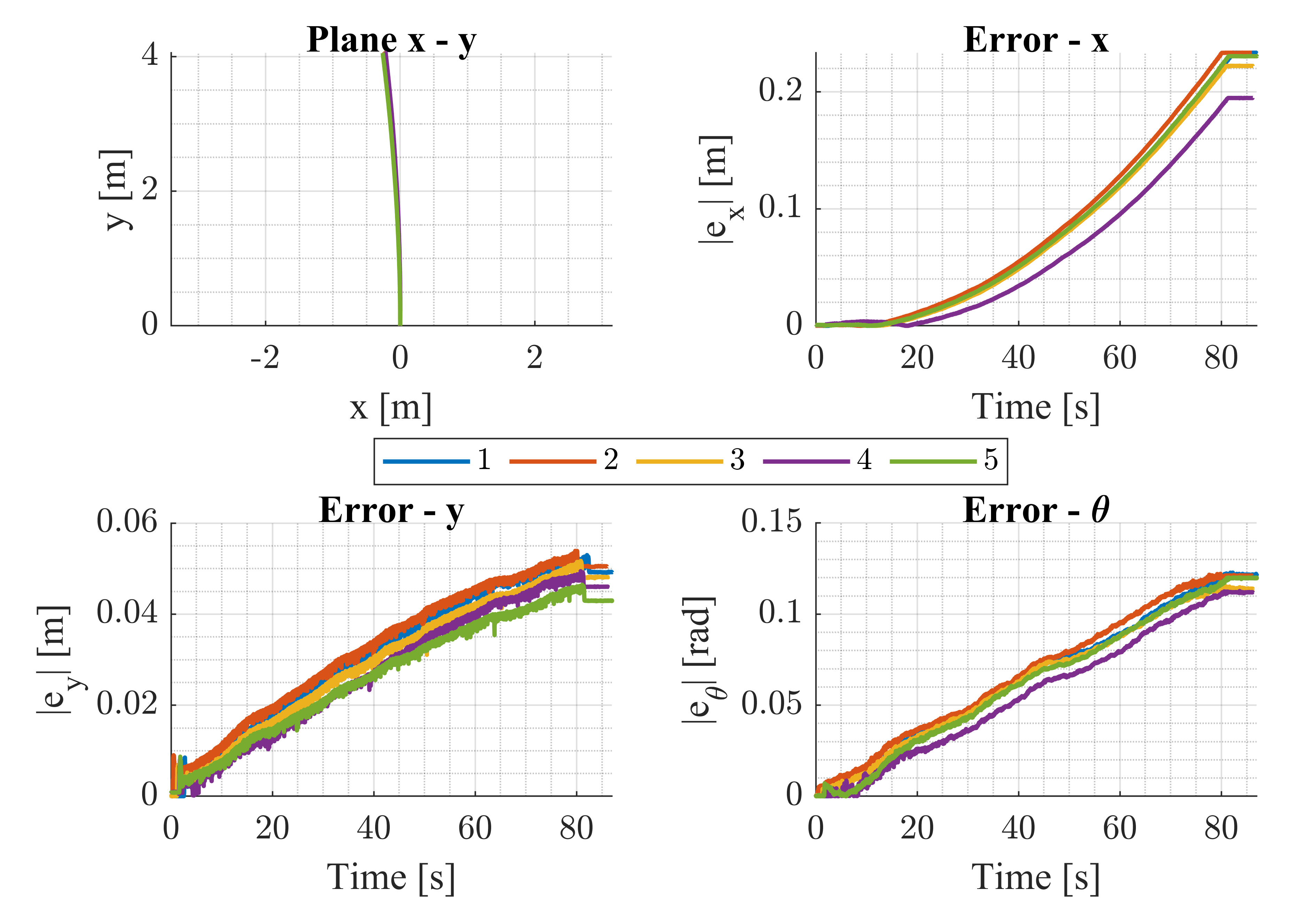}
\caption{Straight line along \textit{x} (x) and \textit{y} axis.}\label{fig_straight_lines}
\end{center}
\end{figure}

\begin{figure}[!ht]
\begin{center}
    \includegraphics[scale=0.50]{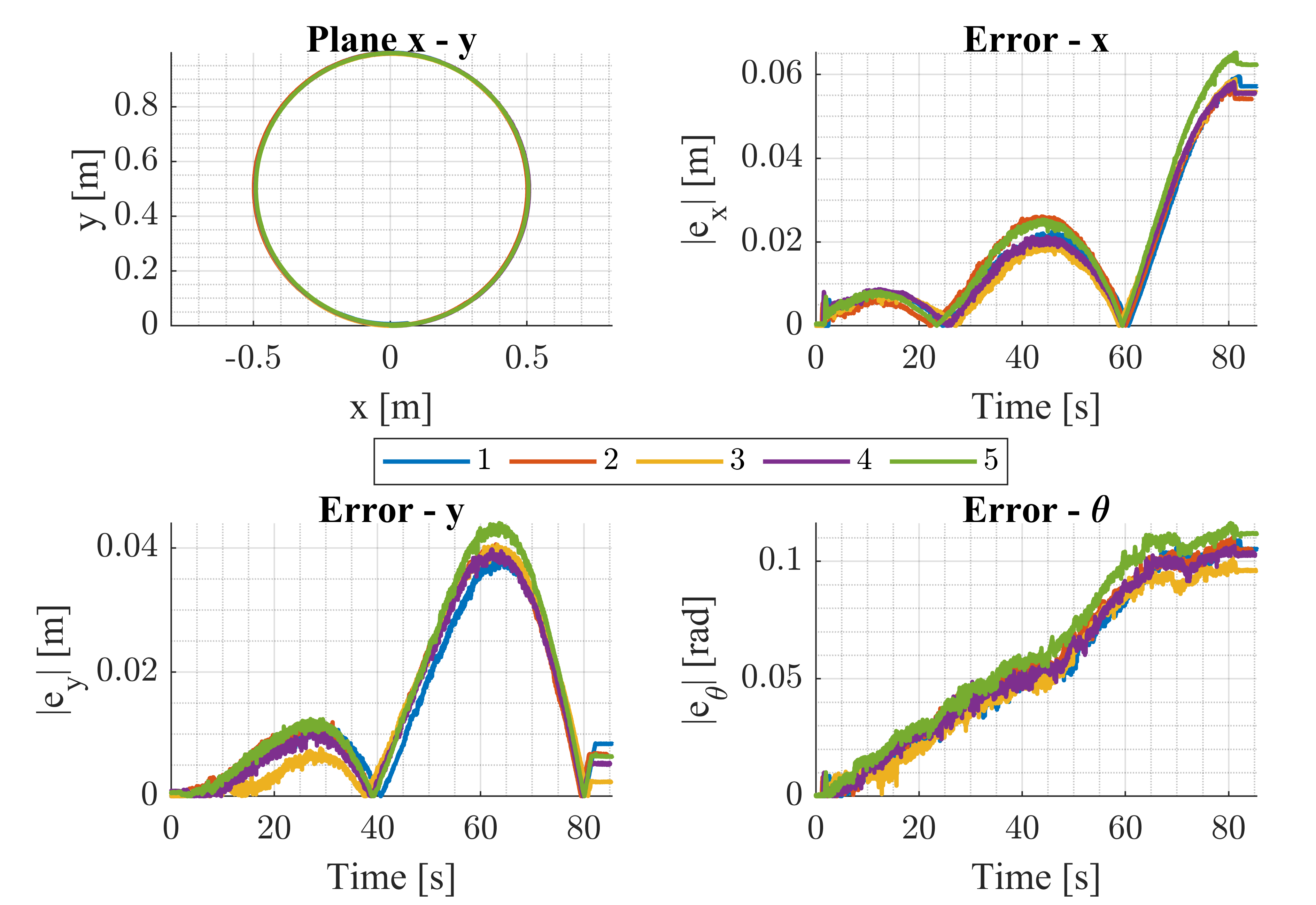} 
\includegraphics[scale=0.50]{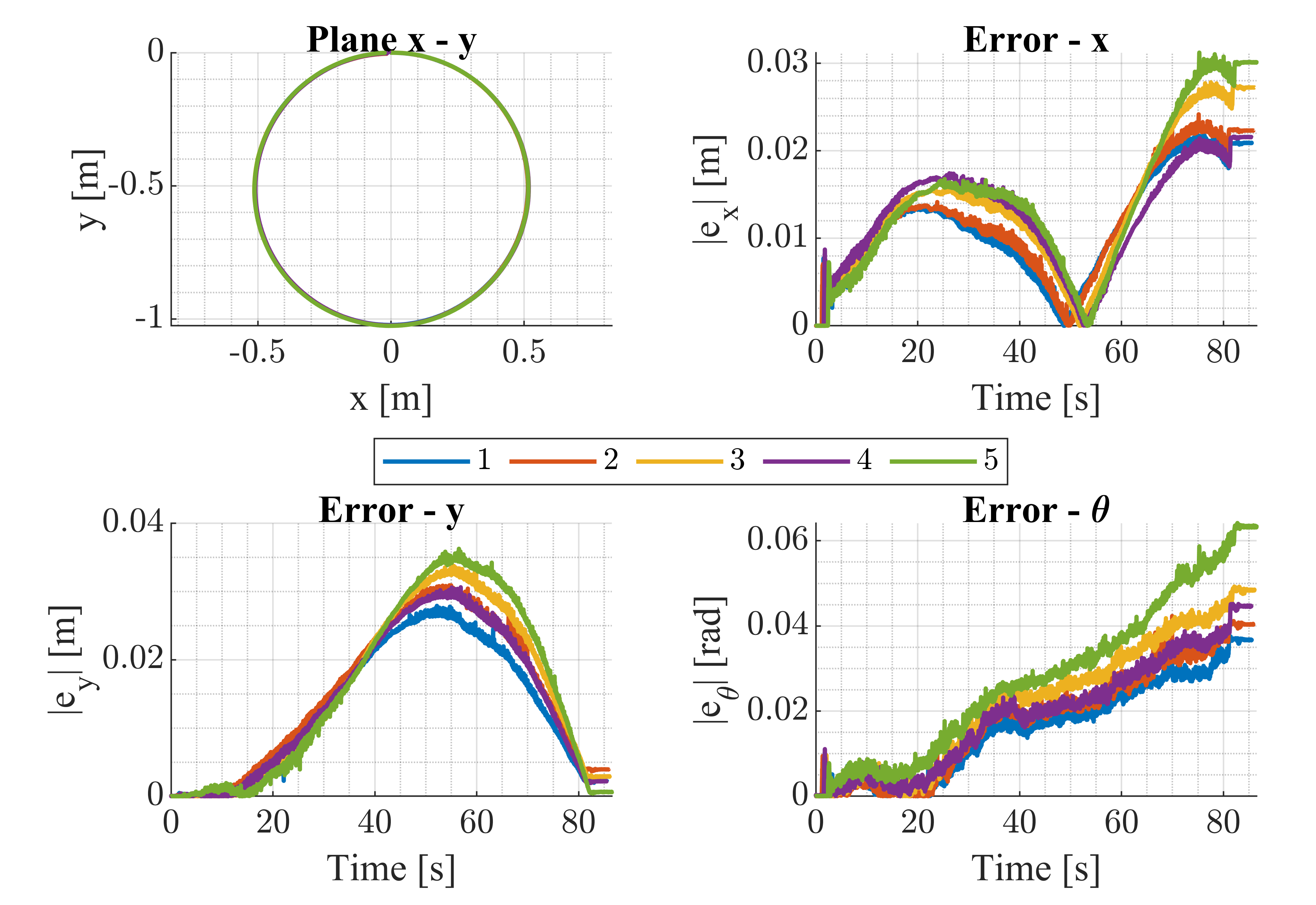}
\caption{Circular movement in CCW (top) and CW (bottom) directions.}\label{fig_circles}
\end{center}
\end{figure}

\begin{figure}[!ht]
\begin{center}
    \includegraphics[scale=0.50]{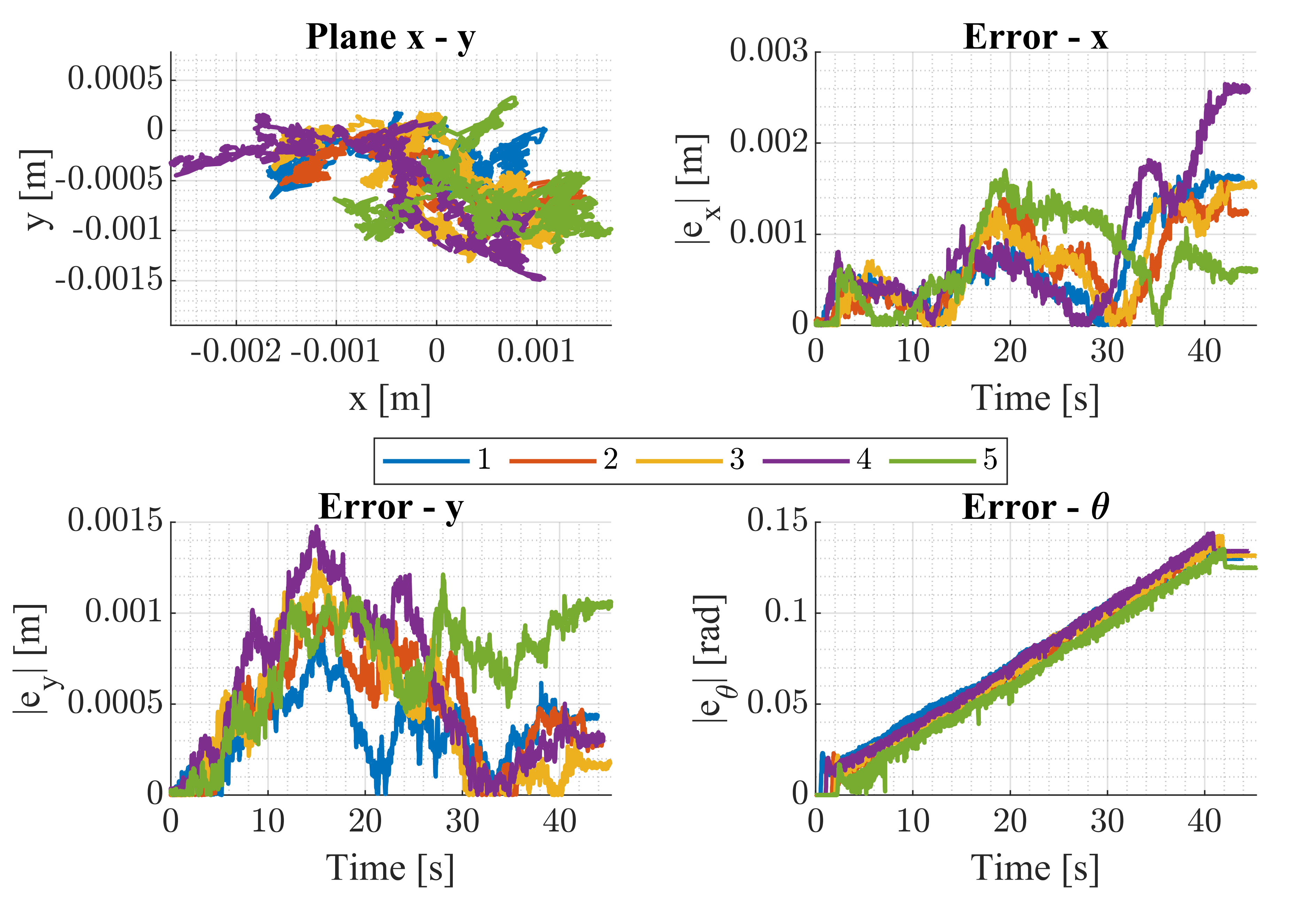} 
\includegraphics[scale=0.50]{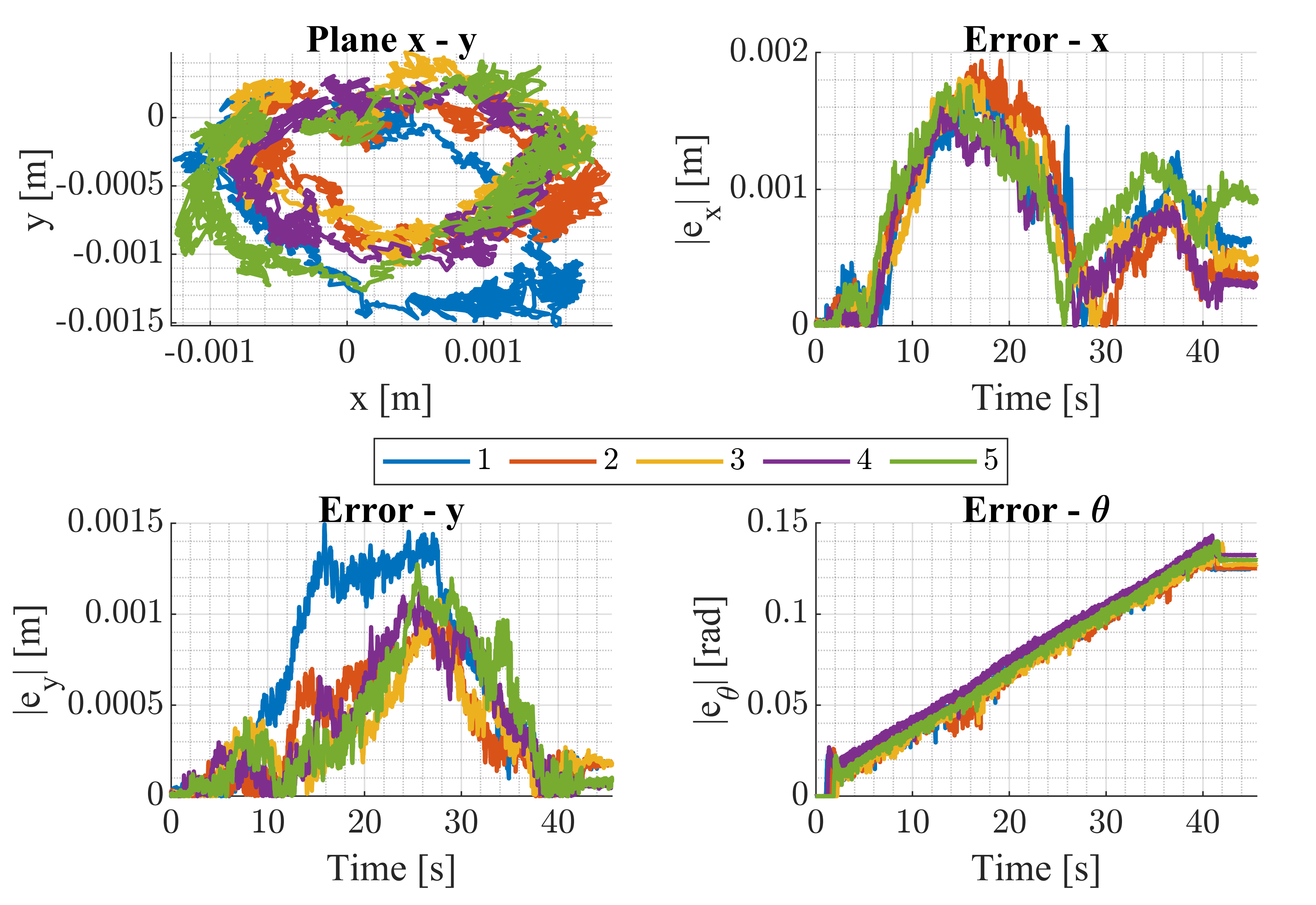}
\caption{Rotational movement around robot's \textit{z} axis in CCW(top) and CW(bottom) direction.}\label{fig_robot_z}
\end{center}
\end{figure}

In this experiment, the robot was driven on horizontal surface with three different types of motion: linear motion, circular motion, and rotation around the robot's $z$ axis. These motions were selected to excite all interactions among the robot's kinematic parameters and to ensure the calibration's validity. During the motion of the robot, data from the motor encoders were recorded and the robot's pose was calculated using kinematic equations and the initial parameters of the robot obtained from its CAD model. In addition to the robot's calculated position, the ground truth position values were recorded using the \textit{OptiTrack} motion capture system to enable comparison and calibration of the robot's actual dimensions. These measures are shown in the upper right corner of Figures \ref{fig_straight_lines}, \ref{fig_circles} and \ref{fig_robot_z}. Linear movement was carried out along the $x$ and the $y$ axis, shown in Figure \ref{fig_straight_lines}.
Circular motions with radius of $0.5\,\mathrm{m}$ and rotation around $z$ axis were carried out in the counterclockwise direction (CCW) and then in the clockwise direction (CW) shown in Figure \ref{fig_circles} and \ref{fig_robot_z} respectively. 
Rotations around the $z$ axis were also performed in counterclockwise and clockwise directions, as shown in Figure \ref{fig_robot_z}.
Furthermore, each of the selected trajectories was executed and recorded five times at low speeds with quintic order polynomial function to ensure smooth velocity and acceleration profiles to minimize the impact of non-systematic errors (slippage and uneven terrain). 

The recordings from the initial tests clearly indicate similar error dynamics, suggesting that the recordings are reliable and that the influence of non-systematic errors has been effectively eliminated. Taking into account six different trajectories with five repetitions, a total of thirty data sets were collected for calibration.

\subsection{Odometry calibration methodology}
Most papers analyzes advancements in systematic odometry calibration, focusing on four mobile robot kinematics: differential drive, Ackerman, tricycle, and omnidirectional \cite{SOUSA2020294} and to the best of author knowledge odometry calibration for 4WIS4WID mobile robot hasn't been done yet. The estimation of a 4WIS4WID mobile robot position and orientation depends on the radii of the wheels $r_i$ and on the distance of each wheel from the center of the robot ($x_{wi}, y_{wi}$). The calibration procedure is formulated as a model fitting problem, since the estimation of the kinematic parameters is performed using a cost function that depends on the error between the ground truth position ($x_{abs}, y_{abs}, \theta_{abs}$) of the mobile robot and position estimated ($x_{est}, y_{est}, \theta_{est}$) using the odometry equation (\ref{eq_kinematics}). The goal of calibration is to find the parameters of the non-linear direct kinematics model that describe the curve with minimal error, i.e. to minimize cost function. 
\begin{equation}
    \begin{split}
     C = &\sum_s\sum_k(x_{est}(s,k) - x_{abs}(s, k))^2
    \\+(y&_{est}(s,k) - y_{abs}(s, k))^2
    + (\theta_{est}(s,k) - \theta_{abs}(s, k))^2.
\end{split}
    \label{eq_cost_function}
\end{equation}

To avoid compensating the total sum of errors with the error sign, the sum of the squared errors for each component of the state vector ($x, y, \theta$) for each discrete step
$k$ and in the $s-th$ recording was chosen.

The vector \textbf{z} is assigned with robot's parameters that needs to be calibrated:
\begin{align}
    \begin{split}
    \mathbf{x}_w^\text{T} &= 
    \begin{bmatrix}
        x_{w1} & x_{w2} & x_{w3} & x_{w4}
    \end{bmatrix}\\
    \mathbf{y}_w^\text{T} &=
    \begin{bmatrix}
        y_{w1} & y_{w2} & y_{w3} & y_{w4}
    \end{bmatrix}\\
    \mathbf{r}_w^\text{T} &=
    \begin{bmatrix}
        r_{1} & r_{2} & r_{3} & r_{4}
    \end{bmatrix}\\
    \mathbf{z}^\text{T} &=
    \begin{bmatrix}
        \mathbf{x}_w^\text{T} & \mathbf{y}_w^\text{T} & \mathbf{r}_w^\text{T}
    \end{bmatrix}.
\end{split}
\end{align}
The parameters must be kept within real limits, which represents the only constraint of the calibration. The optimization problem is defined as
\begin{equation} 
    \min_{\mathbf{z}}C(\mathbf{z}) \quad \text{s.t.} \quad \mathbf{LB} \leq \mathbf{z} \leq \mathbf{UB},
    \label{optimization_problem}
\end{equation}
which minimizes the least squares function (\ref{eq_cost_function}) that has nonlinear dependent on the direct kinematics parameters. The perturbation of the model parameters is constrained by the lower bound (\textbf{LB}) and upper bound (\textbf{UB}) matrices, which limit each parameter to vary within ±5\% of its nominal value. These bounds define the allowable range for the optimization or estimation process, ensuring that the identified parameters remain physically valid and close to the initially assumed (nominal) values.
For parameter estimation, problem statment (\ref{optimization_problem}) is solved using four methods, two gradient - interior-point and Levenberg-Marquardt and two stochastic - Genetic Algortihm and Particle Swarm. All methods were done offline using MATLAB. In general, this involves finding a vector \textbf{z} that minimizes the sum of squares function subject to possible constraints. In this case, the calibration problem has one inequality constraints of upper and lower bounds for the sought vector. It can be concluded that the calibration problem described belongs to the subset of problems that these methods address.

\subsection{Odometry calibration results}
After collecting data for the six trajectories, each repeated five times, the complete dataset was offline optimized for kinematic parameters to find optimal parameters that minimize the error between the ground truth values and the values calculated using (\ref{eq_kinematics}). Subsequently, new data were collected with the parameters obtained and again subjected to offline optimization. With these parameters, the robot's poses were recorded again. This iterative process continued until the results obtained were no longer better than those of the previous recording session. 

\begin{figure}[!ht]
\begin{center}
    \includegraphics[scale=0.50]{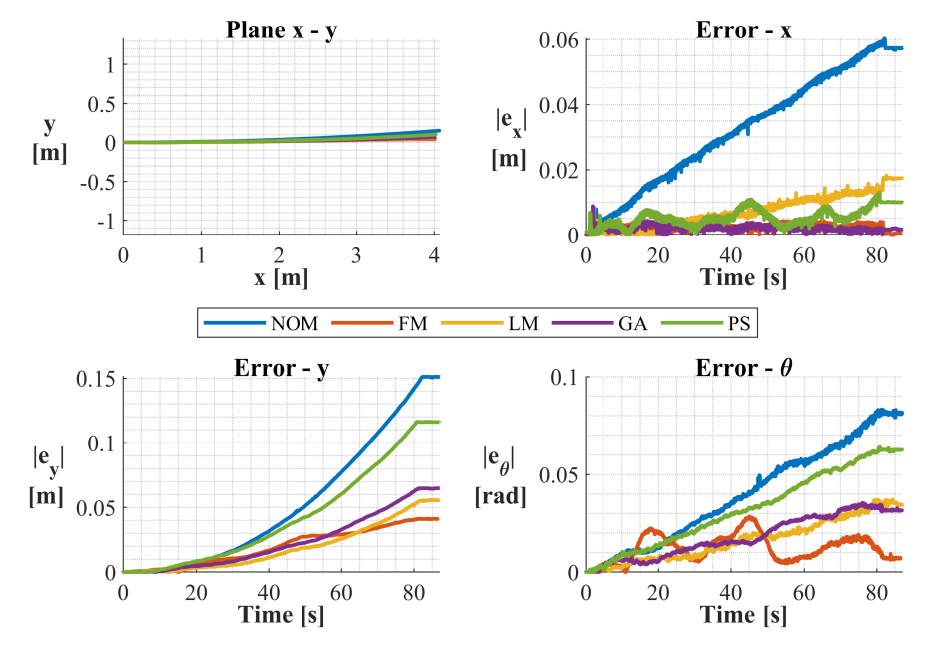} 
\includegraphics[scale=0.50]{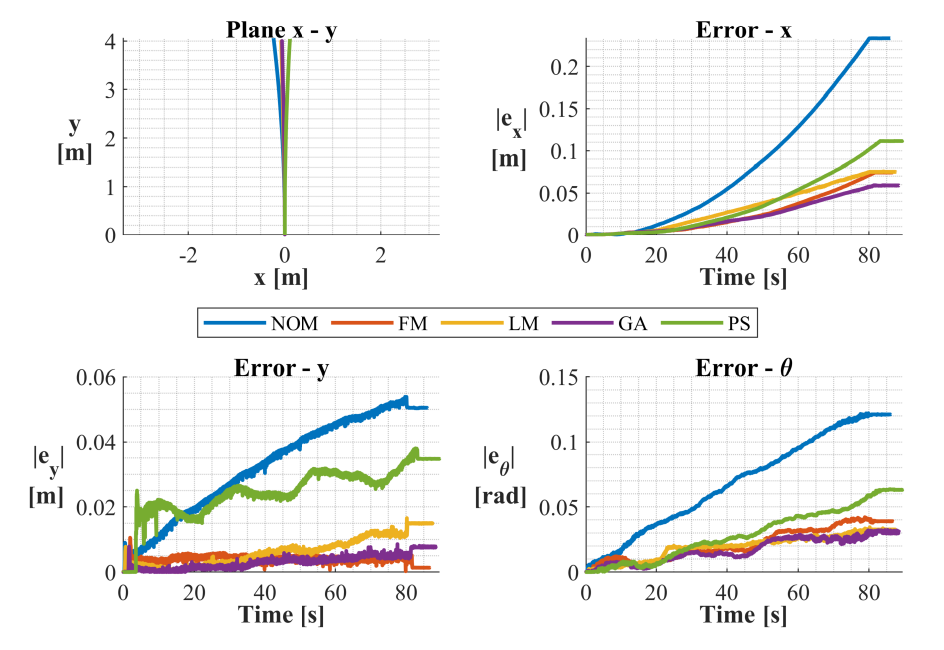}
\caption{Calibration methods comparison for straight line along \textit{x} and \textit{y} axis.}\label{fig_calib_straight_lines}
\end{center}
\end{figure}

\begin{figure}[!ht]
\begin{center}
    \includegraphics[scale=0.50]{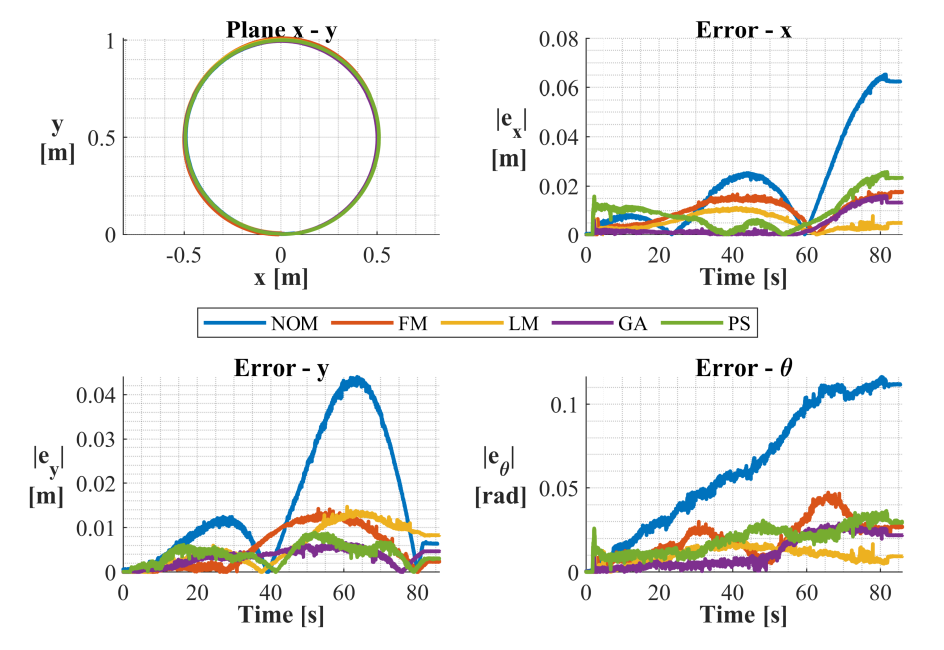} 
\includegraphics[scale=0.50]{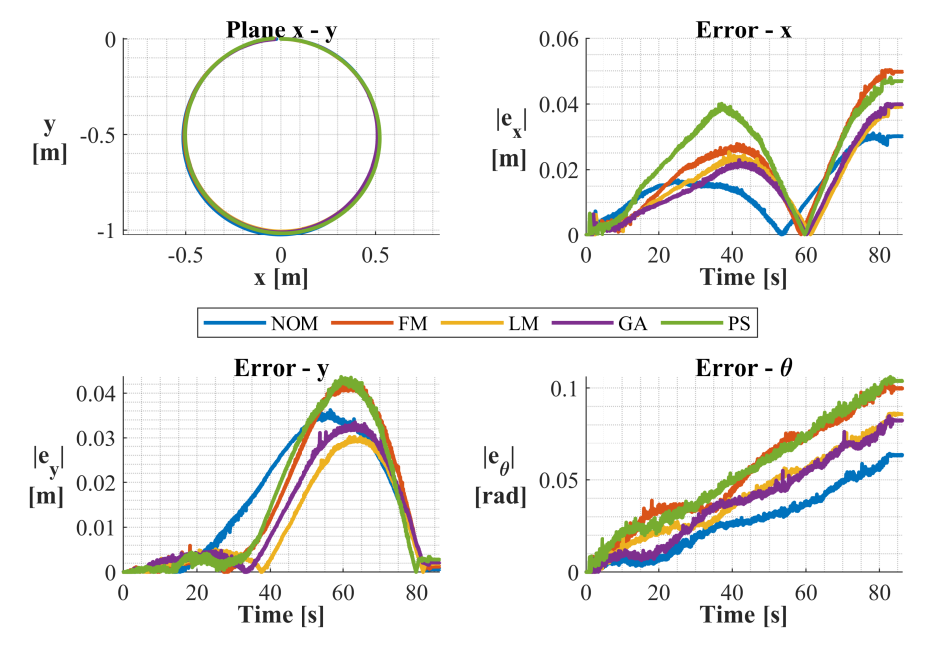}
\caption{Calibration methods comparison for circular movement in clockwise and counterclockwise direction.}\label{fig_calib_circle}
\end{center}
\end{figure}

\begin{figure}[!ht]
\begin{center}
    \includegraphics[scale=0.50]{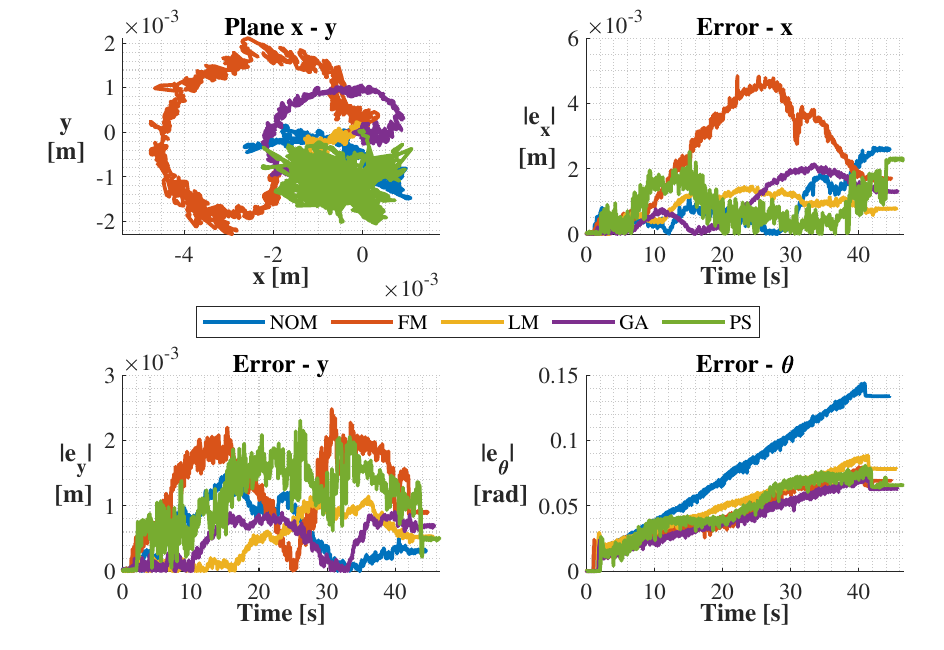} 
\includegraphics[scale=0.50]{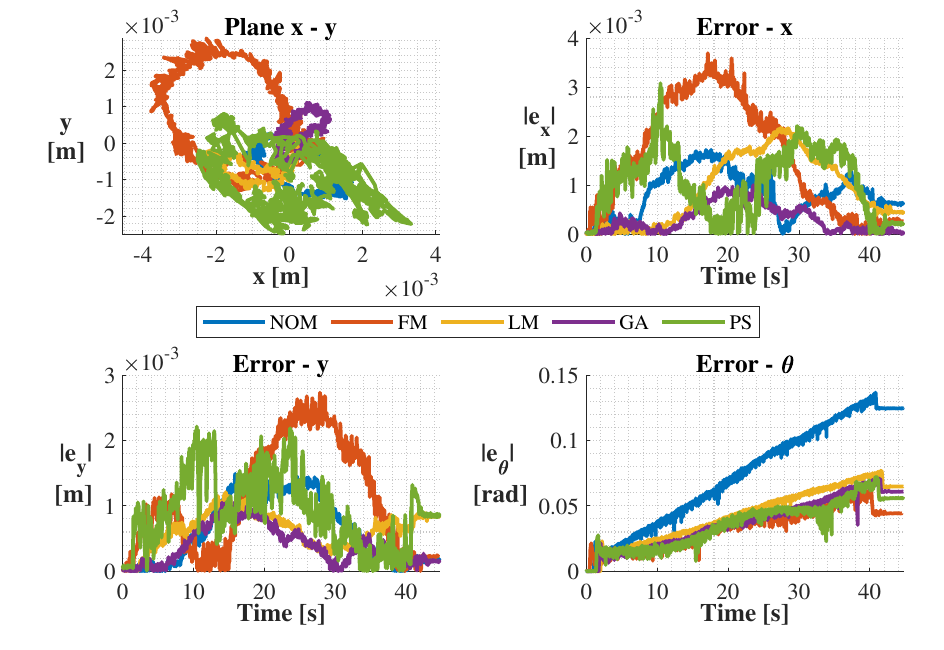}
\caption{Calibration methods comparison for rotational movement around \textit{z} axis in clockwise and counterclockwise direction.}\label{fig_calib_z_axis}
\end{center}
\end{figure}

Figures \ref{fig_calib_straight_lines}, \ref{fig_calib_circle} and \ref{fig_calib_z_axis} graphically illustrate measurements taken before and after calibration. The figures include measurements with nominal parameters obtained from the CAD model(\textit{NOM}) and those obtained using different optimization algorithms: the \textit{fmincon} method with the interior-point algorithm (\textit{FM}), Levenberg-Marquardt optimization (\textit{LM}), a stochastic Genetic Algorithm (\textit{GA}), and the Particle Swarm algorithm (\textit{PA}). Table \ref{table:1} shows the kinematic parameters obtained for the mobile robot associated with these results. The figures and tables \ref{table:2} and \ref{table:3} in Appendix clearly show that the optimized parameters provided significantly improved results.

However, an exception is observed, where performance was worse after calibration, during a counterclockwise circular motion. The reason for this is that all trajectories were used simultaneously during optimization to capture the broadest possible range of robot motion. These discrepancies should not significantly impact the practical robot movements on the wall, as the mobile robot is designed for vertical surfaces primarily utilizes a grid-scan path for performing NDT, which mostly employs linear type of motion.
\begin{table}[!ht]
\centering
\caption{Nominal and calibrated odometry parameters}
\label{table:1}
\scriptsize
\scalebox{0.66}{
\begin{tabular}{ |c|c|c|c|c|c|c|c|c|c|c|c|c| } 
 \hline
 & $x_{w1}$ & $x_{w2}$ & $x_{w3}$ & $x_{w4}$ &  $y_{w1}$ & $x_{y2}$ & $y_{w3}$ & $x_{y4}$ &
  $r_{1}$ & $x_{2}$ & $r_{3}$ & $r_{4}$\\ \hline
 NOM  & 112,5 & -112,5 & -112,5 & 112,5 & 112,5 & 112,5 & -112,5 & -112,5 & 25,4 & 25,4 & 25,4 & 25,4\\ 
 \hline
 FM & 118,1 & -118,1 & -116,4 & 112,4 & 106,9 & 118,1 & -107,1 & -106,9 & 25,55 & 26,67 & 25,91 & 25,11 \\ 
 \hline
 LM & 114,7 & -114,7 & -113,2 & 113,2 & 114,2 & 114,2 & -113,9 & -107,7 & 25,83 & -25,86 & 25,86 & 25,36\\
 \hline
 GA & 111,4 & -114,0 & -112,2 & 107,4 & 114,4 & 114,7 & -116,3 & -116,2 & 26,04 & 25,93 & 25,67 & 25,46\\
 \hline
 PS & 107,9 & -116,5 & -117,5 & 108,1 & 114,4 & 111,7 & -116,7 & -108,5 & 26,11 & 25,13 & 26,61 & 25,01\\
 \hline
\end{tabular}}

\end{table}

\section{POSE ESTIMATOR DESIGN} \label{sec_pose_estimator}
\begin{figure}[!ht] 
    \centering
    \includegraphics[scale=0.53]{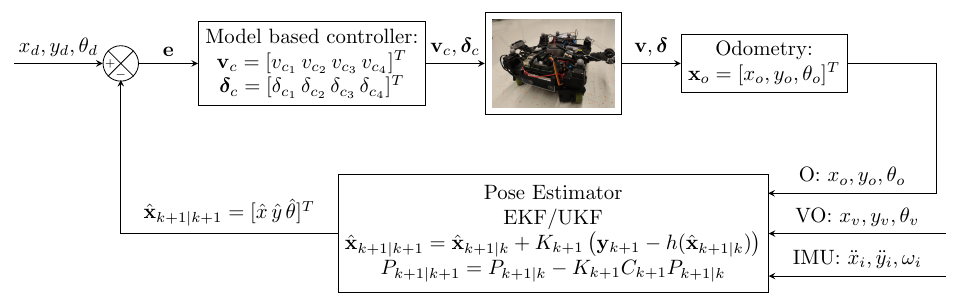}
    \caption{Wall climbing robot control systems design.}\label{fig_wcr_cs_design}
\end{figure}

Due to its extensive applicability within mobile robotics, the Kalman filter has attracted significant attention from researchers \cite{URREA20211}. State estimators for mobile robots are often based on the application of perception sensors, such as LIDAR sensors and 3D cameras, which enable mapping and localization within the environment. In the case of mobile robots operating on vertical surfaces, it is often impossible to use such sensors, necessitating the use of robot odometry, IMU, and visual odometry to estimate the robot's position on the wall, thereby improving the accuracy of and performances of such autonomous system. Figure \ref{fig_wcr_cs_design} illustrates a schematic diagram of the control law, the robot, and the proposed state estimator, which will be based on the sensor fusion with Extended Kalman Filter (EKF) and Unscented Kalman Filter (UKF).
For the following trajectory, a non-linear model-based controller was designed \cite{CARAN2024} and optimal control-based controller \cite{CARAN2024ACTUATORS}, where input to the controller is error $\mathbf{e} = [x_e \quad y_e \quad \theta_e]^\text{T}$ between estimated pose from the pose estimator $\hat{\mathbf{x}}_{k+1|k+1}$ and desired trajectory $\mathbf{x}_d=[x_d \quad y_d \quad \theta_d]^\text{T}$, outputs of the controller are motors velocities $\mathbf{v}_c$ and the steering angles $\bm{\delta}_c$.
EKF and UKF are frequently utilized filters for nonlinear models of mobile robots to estimate the robot's position and orientation \cite{DALFONSO2015122}. Nonlinear filtering is the problem of estimating the state of
a nonlinear stochastic system from noisy measurements. For discrete-time systems such framework is represented by the equations 
\begin{equation}
\begin{aligned}
    x_{k+1} &= f(x_k, u_k) + w_k\\
    z_k &= h(x_k) + v_k,
\end{aligned}
\end{equation}
where $x$ is state to be estimated, $u = \begin{bmatrix}
    \mathbf{v}_c & \bm{\delta}_c
\end{bmatrix}^\text{T}$is the control input and $w_k$ and $v_k$ represents system and measurements model. Model from (\ref{eq_kinematics}) belongs this framework.
In the subsequent section, the design of a state estimator based on EKF and UKF for a 4WIS4WID mobile robot will be described and the results will be presented throughout the remainder of the chapter.

\subsection{Extended Kalman filter}
The Extended Kalman Filter has been used for many years to estimate the states of a nonlinear system by fusing multiple noisy measurements. It is based on linearization of the nonlinear functions ($f, h$) around fixed point and on assumption that the initial state and  measurment noises are not correlated and can be modeld using Gaussian. From application side, EKF is simply a time-varying Kalman filter where dynamic and measurement matrices are linearized by:
\begin{equation}
    A_k = \left. \frac{\partial f(x, u)}{\partial x} \right|_{\substack{x = \hat{\mathbf{x}}_{k|k} \\ u = u_k}} \\
    C_k = \left. \frac{\partial h(u)}{\partial x} \right|_{\substack{x = \hat{\mathbf{x}}_{k|k-1}}}
\end{equation}
and the output is $\hat{x}_{k|k}$ and matrix $P_{k|k}$ starting from the given initial states $\hat{x}_{0|0}$ and $\hat{P}_{0|0}$ the algorithm below \cite{DALFONSO2015122}

\subsection{Unscented Kalman filter}
The main idea  of the Unscented Kalman Filter (UKF) involves identifying a transformation that enables the approximation of the mean and covariance of an $n$-dimensional random vector when it undergoes a nonlinear transformation \cite{JULIER2004401}. This is achieved by generating a set of $2n + 1$, called $\sigma$ points based on the mean and covariance of the original vector. These sigma points are then passed through the nonlinear function, and the mean and covariance of the transformed vector are approximated using the transformed $\sigma$ points. 
Regarding its approximation capabilities, it has been shown that while the Extended Kalman Filter (EKF) provides a first-order accurate state estimate, the UKF achieves third-order accuracy when the system is subject to Gaussian noise. 

Additionally, the EKF approximates the covariance to first-order accuracy, whereas the UKF does so to second-order accuracy. The estimated state  are computed in the same way as in EKF, the only difference is in how the Kalman gain is calculated. The reader is referred to \cite{WAN2000} and \cite{Wu2005357}  for further theoretical details.
\section{EXPERIMENTAL VALIDATION}\label{sec_experimental_validation}
\begin{figure}[!ht]
\begin{center}
    \includegraphics[scale=0.25]{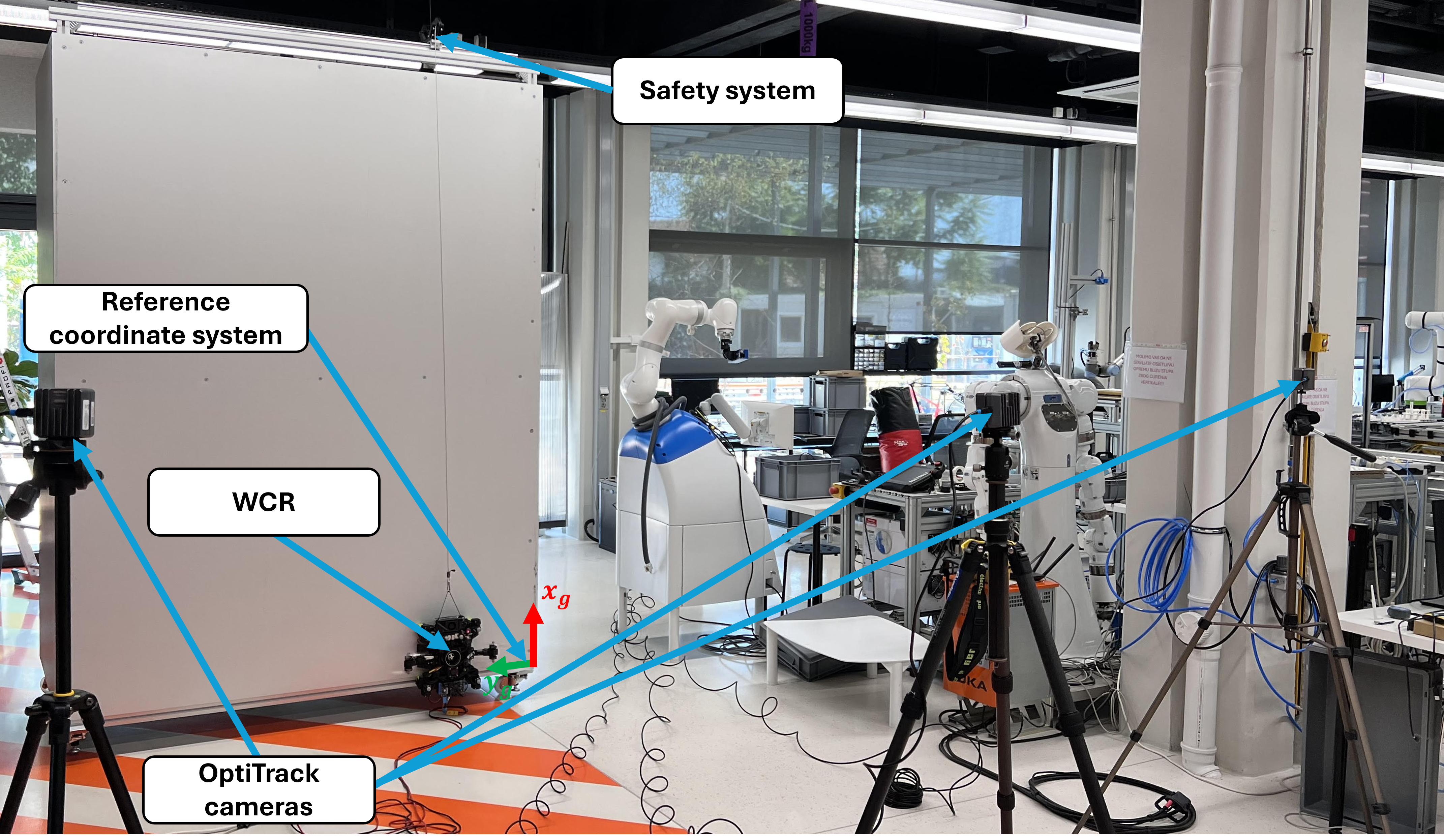} 
\caption{Experimental setup for pose estimator comparison.}\label{odom_ekf_ukf_setup}
\end{center}
\end{figure}

The trajectory tracking along the $X$ and $Y$ axes was compared on real system, where trajectory tracking using odometry, EKF, and UKF was evaluated. A wall equipped with a safety system at the top was used to prevent the robot from falling and getting damaged in case of a failure. The surface used was melamine faced chipboard, which has significantly lower friction than concrete, in order to include more critical scenarios in the experiment. Figure \ref{odom_ekf_ukf_setup} shows the experimental setup used for state estimator validation. The global coordinate system is defined such that the positive $X$ axis direction extends from the bottom to the top of the wall, while the positive $Y$ axis direction is oriented from right to left.

\begin{figure}[!ht]
\begin{center}
    \includegraphics[scale=0.6]{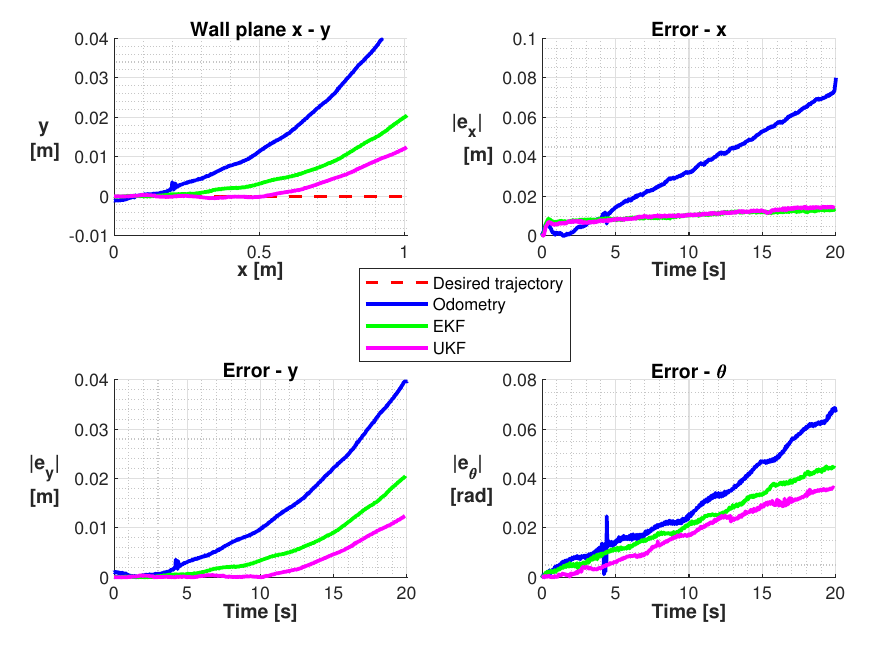} 
\caption{Experimental results for pose estimators in $X$ axis direction.}\label{x_axis_odom_ekf_ukf}
\end{center}
\end{figure}

\begin{figure}[!ht]
\begin{center}
    \includegraphics[scale=0.6]{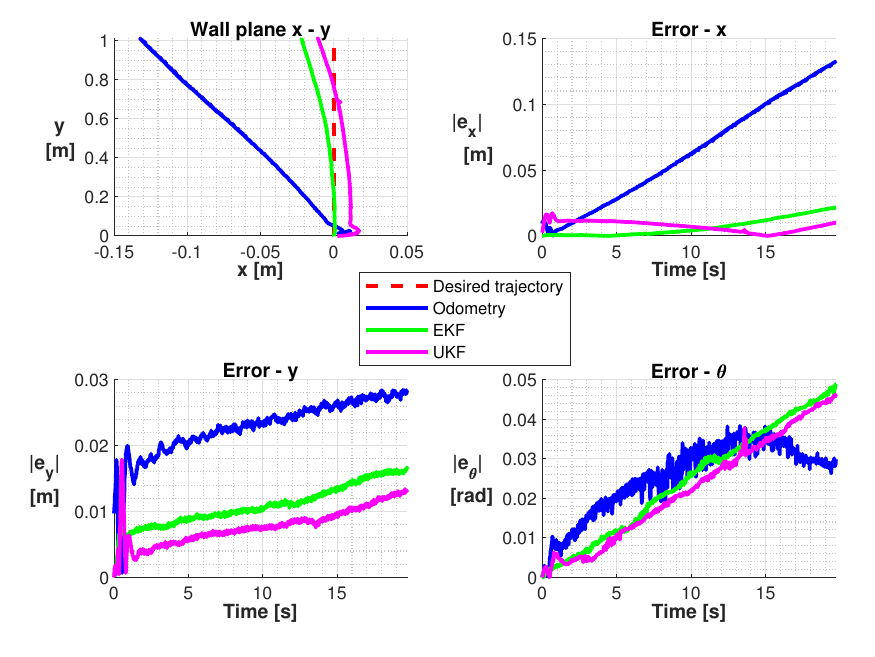} 
\caption{Experimental results for pose estimators in $Y$ axis direction.}\label{y_axis_odom_ekf_ukf_}
\end{center}
\end{figure}

The results presented in the Figures \ref{x_axis_odom_ekf_ukf} and \ref{y_axis_odom_ekf_ukf_} demonstrate that both the EKF and UKF provide superior performance compared to odometry alone, particularly in orientation estimation. This improvement is primarily due to the incorporation of the IMU, which effectively detects deviations in the robot’s orientation. Furthermore, it is evident that relying solely on odometry leads to slippage. During linear motion along the $X$ axis, the robot is unable to reach the intended endpoint without the assistance of a state estimator, as a result of slippage. A similar issue arises during linear motion along the $Y$ axis, where the robot exhibits unintended displacement along the $X$ axis, attributed to gravitational effects. The integration of EKF and UKF, in conjunction with IMU and visual odometry, successfully mitigates these issues. This is achieved by enabling the estimation of gravitational influences and compensating for them by increasing the robot's velocity, thus improving the trajectory tracking performance. Parameters used for UKF are: $\alpha=0,001$, $\beta=2.0$, $\kappa=0$.

\section{CONCLUSION} \label{sec_conclusion}
In this paper, the design and implementation of a pose estimation system for a 4WIS4WID mobile wall climbing robot have been developed and validated through experimental testing. The kinematic model of the robot was first formulated, and a systematic odometry calibration procedure was applied using both gradient-based and stochastic optimization methods to minimize pose estimation errors. A state estimation framework fusing odometry, visual odometry, and inertial measurements was designed using EKF and UKF. The proposed estimation algorithms were integrated into the real system and tested on a wall climbing robot. Experimental results demonstrated that both EKF and UKF significantly improved pose estimation accuracy compared to odometry alone, especially in reducing orientation errors and compensating for slippage and gravitational effects. The developed system enables reliable pose estimation necessary for precise navigation and execution of non-destructive testing tasks on vertical surfaces. Further research will focus on implementing more robust sensor fusion techniques, investigating the estimator’s performance under different surface conditions, as well as including more accurate dynamic model of the robot with friction model between wheels and surface.

\section*{ACKNOWLEDGMENT}

The authors would like to acknowledge the support of the Regional Centre of Excellence for Robotic Technology (CRTA), funded by the ERDF fund. Additionally, we would like to thank our student Jakov Vitko for his assistance in collecting the data for the calibration.

\section*{APPENDIX}
Tables~II and III present errors of the robot’s motion from the reference trajectory with nominal parameters in table I and calibrated parameters in table II, where
$e_{x,m}$, $e_{y,m}$, and $e_{\theta,m}$ denote the maximum errors in $x$, $y$, and $\theta$, respectively,
and $e_{x,a}$, $e_{y,a}$, and $e_{\theta, a}$ denote the corresponding average errors.
\begin{table}[!ht]
\centering
\caption{ODOMETRY ERROR BEFORE CALIBRATION}
\label{table:2}
\scriptsize
\scalebox{0.66}{
\begin{tabular}{ |c|c|c|c|c|c|c|c|c|c|c|c|c| } 
 \hline
 & $e_{x,m}$ & $e_{x,a}$ & $e_{y,m}$ & $e_{y,a}$ &  $e_{\theta,m}$ & $e_{\theta,a}$\\
 \hline
 Line X & 0,063 & 0,062 & 0,152 & 0,14 & 0,099 & 0,081\\ 
 \hline
 Line Y & 0.234 & 0.223 & 0,054 & 0,051 & 0,123 & 0,119\\ 
 \hline
 Circle CCW & 0,065 & 0,06 & 0,044 & 0,041 & 0,116 & 0,109\\
 \hline
 Circle CW & 0,091 & 0,025 & 0,036 & 0,032 & 0,064 & 0,048\\
 \hline
 Angular CCW & 0,003 & 0,002 & 0,001 & 0,001 & 0,144 & 0,141\\
 \hline
 Angular CW & 0,002 & 0,002 & 0,001 & 0,001 & 0,143 & 0,0139\\
 \hline
\end{tabular}}
\end{table}

\begin{table}[!ht]
\centering
\caption{ODOMETRY ERROR AFTER CALIBRATION}
\label{table:3}
\scriptsize
\scalebox{0.66}{
\begin{tabular}{ |c|c|c|c|c|c|c|c|c|c|c|c|c| } 
 \hline
 & $e_{x,m}$ & $e_{x,a}$ & $e_{y,m}$ & $e_{y,a}$ &  $e_{\theta,m}$ & $e_{\theta,a}$\\
 \hline
 Line X(FM) & 0,008 & 0,008 & 0,152 & 0,14 & 0,099 & 0,081\\ 
 \hline
 Line Y(GA) & 0,059 & 0,036 & 0,009 & 0,008 & 0,032 & 0,025\\ 
 \hline
 Circle CCW(LM) & 0,065 & 0,06 & 0,044 & 0,041 & 0,116 & 0,109\\
 \hline
 Circle CW(LM) & 0,011 & 0,009 & 0,015 & 0,009 & 0,019 & 0,013\\
 \hline
 Angular CCW(LM) & 0,002 & 0,002 & 0,001 & 0,001 & 0,075 & 0,07\\
 \hline
 Angular CW(LM) & 0,002 & 0,002 & 0,001 & 0,001 & 0,08 & 0,075\\
 \hline
\end{tabular}}
\end{table}

\nocite{*}

\bibliographystyle{IEEEtran}
\bibliography{references}

\end{document}